# CCC-Guided Latent Representation Learning Enhances Cell-Cell Communication Analysis in Single-Cell RNA-seq Data

Cong Qi
Department of Computer Science
New Jersey Institute of Technology

Yeqing Chen
Department of Computer Science
New Jersey Institute of Technology

Zhi Wei
Department of Computer Science
New Jersey Institute of Technology

## Abstract

Cell–cell communication (CCC) plays a central role in coordinating biological functions across cell types and states. While recent computational tools leverage single-cell RNA sequencing (scRNA-seq) data to infer CCC networks via ligand–receptor interactions, they typically rely on predefined cell clusters, which may introduce biases or overlook important signaling dynamics. In this work, we present CCCVAE, a novel variational autoencoder framework that integrates CCC information into the representation learning process. Rather than treating clustering and communication inference as separate tasks, CCCVAE incorporates intercellular signaling strength into the latent prior distribution, guiding the model to learn embeddings that reflect both transcriptomic similarity and functional interaction. Our approach enables more accurate clustering and enhanced detection of biologically meaningful communication pathways. We validate CCCVAE across multiple single-cell datasets and show that it consistently improves latent structure discovery and communication network inference compared to state-of-the-art baselines. Our results demonstrate that CCC can serve not only as a target of analysis but also as a powerful supervisory signal for single-cell representation learning.

## 1 Introduction

Cell–cell communication (CCC) is fundamental to the organization and function of multicellular systems. It governs a wide array of biological processes, including development, immune response, and tissue homeostasis, by enabling cells to transmit and receive biochemical signals. Traditionally, studies of CCC were limited to a few cell types and marker genes due to experimental constraints. The advent of single-cell RNA sequencing (scRNA-seq) has revolutionized this landscape by enabling transcriptomic profiling at single-cell resolution, thus providing an unprecedented opportunity to infer intercellular communication across diverse cell populations in complex tissues [2, 40]. A growing number of computational methods [42, 18, 5, 7, 27, 22] now aim to reconstruct CCC networks by quantifying ligand–receptor interactions between cell populations using scRNA-seq data. These approaches, including CellChat [18], CellPhoneDB [7], and NicheNet [5], typically begin by clustering cells into discrete groups—often representing putative cell types or states—and then estimate communication scores based on the co-expression of known ligand and receptor gene pairs between clusters. These inferred interactions are used to construct cell–cell signaling networks that reflect the putative modes of communication between defined cell groups. Figure 1 shows the illustration of cell-cell communication.

While effective in many contexts, these methods are inherently limited by a critical assumption: that the clustering of cells is accurate and biologically meaningful. In reality, clustering results may be sensitive to algorithmic choices, preprocessing steps, or biological noise in the data [43, 30]. If the underlying clustering is suboptimal—merging distinct cell types or splitting a homogeneous population arbitrarily—then downstream CCC inference may be misleading. For example, if two subpopulations are erroneously grouped into a single cluster, any potential communication between them may be missed entirely. Conversely, if one cell type is split into multiple clusters, artificial intra-type communication may be inferred, which lacks biological validity.

This limitation is particularly consequential because inferred CCC networks often guide downstream hypotheses and experimental validations. Thus, improving the fidelity of CCC inference depends not only on better modeling of ligand–receptor activity but also on recognizing and addressing the limitations of cell clustering itself. Despite this, most existing CCC tools treat clustering as a preprocessing step that is decoupled from communication inference. This disconnect motivates the need for integrative models that consider CCC information during the process of learning cell representations and forming clusters [2, 39].

We propose a new framework to bridge this gap by integrating CCC inference with representation learning and clustering, thereby aligning the analytical goals of cell-type discovery and communication analysis. Our key hypothesis is that CCC information, particularly ligand–receptor pair activity between cells, can serve as a valuable supervisory signal to guide the learning of latent cell representations. If two cells exhibit strong communication via ligand–receptor interactions, it may be preferable to assign them to different clusters to accurately model the directionality and specificity of signaling.

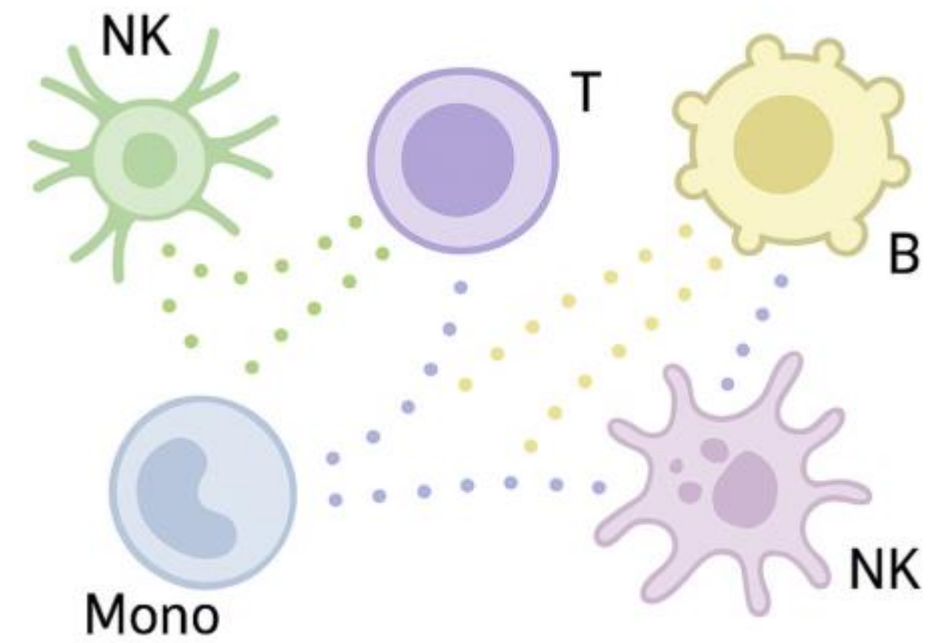

Figure 1: Schematic representation of intercellular communication among cell types. Each labeled cell type—NK cells, T cells, B cells, and monocytes—interacts through ligandreceptor signaling, illustrated as colored dots. Ligands secreted by one cell bind to receptors on another, enabling immune coordination. Quantifying the expression levels of ligand-receptor pairs facilitates the study of immune cell communication dynamics in peripheral blood.

To formalize this idea, we introduce a novel latent representation learning approach based on a variational autoencoder (VAE) framework. Our method builds upon the variational autoencoder (VAE) framework [34, 1]. Unlike traditional VAEs that assume independence in the prior distribution, we incorporate CCC information by modifying the prior to reflect intercellular communication patterns. Our model, termed CCCVAE, incorporates CCC-derived constraints through the latent prior distribution. While existing methods rely on fixed clusters, CCCVAE dynamically refines representation learning based on communication evidence. This enables more accurate identification of both cell types and the signaling interactions that define their relationships. We demonstrate that our approach improves downstream communication inference and yields more interpretable groups.

## 2 Related work

### 2.1 Variational Autoencoders for Single-Cell transcriptomics

Unsupervised representation learning has become essential in decoding the complexity of single-cell RNA sequencing (scRNA-seq) data. Among various approaches, Variational Autoencoders (VAEs) [20, 31, 15] offer a generative framework that is both flexible and probabilistic, enabling the modeling of noisy, high-dimensional count data [21, 33]. In the single-cell setting, the generative process typically assumes that each cell $x_n$ arises from a low-dimensional latent variable $z_n$ via:

$$z_n \sim \mathcal{N}(0, I), x_n \sim p_\theta(x_n \mid z_n), \text{ and } p_\theta(x_n \mid z_n) = \prod_{g=1}^{G} NB\left(x_{ng} \mid \mu_{ng}(z_n), \theta_g\right)$$

Here, each gene expression count $x_{ng}$ is modeled using a Negative Binomial distribution with genespecific dispersion $\theta_g$, and the mean $\mu_{ng}$ is parameterized by a neural decoder $f_\theta(z_n)$. The model is trained by maximizing the Evidence Lower Bound (ELBO), which approximates the intractable marginal likelihood:

$$\mathcal{L}_{ELBO} = \mathbb{E}_{q_\phi(z_n|x_n)}[\log\, p_\theta(x_n \mid z_n)] - KL\left(q_\phi(z_n \mid x_n) \| p(z_n)\right)$$

This objective balances two competing terms: the reconstruction likelihood of observed data under the generative model and the Kullback-Leibler divergence between the approximate posterior and the standard normal prior over the latent space. Extensions of this framework, such as scVAE [14] , LDVAE [10], DCA [8], and scGen [24], have explored modifications to the decoder architecture or latent prior to improve interpretability and capture dynamic cellular processes. Notably, scVI [23] introduced a hierarchical Bayesian model tailored to scRNA-seq, incorporating additional latent variables for sequencing depth and using variational inference to model gene expression with negative binomial likelihoods. Building upon these foundations, siVAE [6] enhances interpretability by designing a VAE that jointly learns sample and gene embeddings, facilitating the identification of gene modules and hubs without explicit network inference. ScInfoVAE [28] employs mutual information regularization within the VAE framework, coupled with a zero-inflated negative binomial model, to effectively capture both local and global structures in scRNA-seq data, thereby improving clustering performance. Furthermore, scVIC [44] addresses biological heterogeneity and batch effects by integrating variational inference with mechanisms to handle technical variability, offering a robust approach for modeling complex single-cell datasets. Despite these advances, most VAE-based models assume that cells are conditionally independent given latent variables, ignoring biologically meaningful dependencies such as intercellular communication or spatial relationships. [37, 3, 29, 38]

### 2.2 Computational modeling of Cell-Cell Communication

Cell-cell communication (CCC), mediated by ligand-receptor interactions, plays a vital role in shaping cell fate, function, and tissue structure. Recent tools such as CellChat [18], CellPhoneDB [7], and NicheNet [5] have demonstrated how CCC can be inferred from scRNA-seq data by leveraging curated ligand-receptor interaction databases [47, 25, 45]. These methods typically evaluate the potential communication between cell groups by scoring co-expression of ligand-receptor pairs, either statistically or with regulatory priors, and have uncovered key signaling mechanisms in development, immunity, and cancer. However, these CCC inference tools are applied in a post hoc manner, detached from the representation learning stage. As a result, learned latent representations from VAEs do not reflect communication-driven dependencies among cells. Although a few models have attempted to incorporate spatial or relational information through graph-based encodings (e.g., spatial-VAE or contrastive approaches), they generally lack a biologically grounded kernel for intercellular communication.[35, 16, 46]

### 2.3 Bridging the gap

To the best of our knowledge, no prior work has incorporated CCC into the representation learning process itself. We propose CCCVAE, the first VAE-based framework that learns a latent space explicitly shaped by intercellular communication. We embed CCC structure into the latent space using a communication-aware Gaussian Process (GP) prior, whose kernel is computed from predicted ligandreceptor activity. This design makes a conceptual shift: we no longer treat clustering or latent space structure as an unsupervised objective, but as a biologically guided process informed by communication topology.

## 3 Method

We aim to learn low-dimensional representations of single-cell gene expression data that not only preserve biological structure but also reflect meaningful cell-cell communication (CCC) patterns. To this end, we propose CCCVAE, a novel variational autoencoder that introduces a CCC-aware Gaussian Process (GP) prior over a subset of the latent dimensions. This approach enables a balance between biologically meaningful communication patterns and flexible latent modeling. Figure 2 shows the framework of our CCCVAE model.

### 3.1 VAE Framework for Single-Cell Data

Let $X = [x_1,...,x_N] \in \mathrm{R}^{N\times G}$ denote the gene expression matrix for $N$ cells and $G$ genes. In CCCVAE, we learn latent representations $z_i \in \mathrm{R}^D$ for each cell $i$, where $D$ is the total latent dimensionality.

We partition the latent space as:

$$z_i = \begin{bmatrix} z_i^{(1)} \\ z_i^{(2)} \end{bmatrix}, \quad z_i^{(1)} \in \mathbb{R}^l, \quad z_i^{(2)} \in \mathbb{R}^{D-l} \tag{1}$$

where:

- $z_i^{(1)}$ (the first $l$ dimensions) encode communication-driven structure governed by a CCC-aware GP prior,
- $z_i^{(2)}$ (the remaining $D - l$ dimensions) model unconstrained variation using a standard Gaussian prior.

This separation allows CCC-relevant information to be structured while maintaining flexibility in modeling residual biological variation.

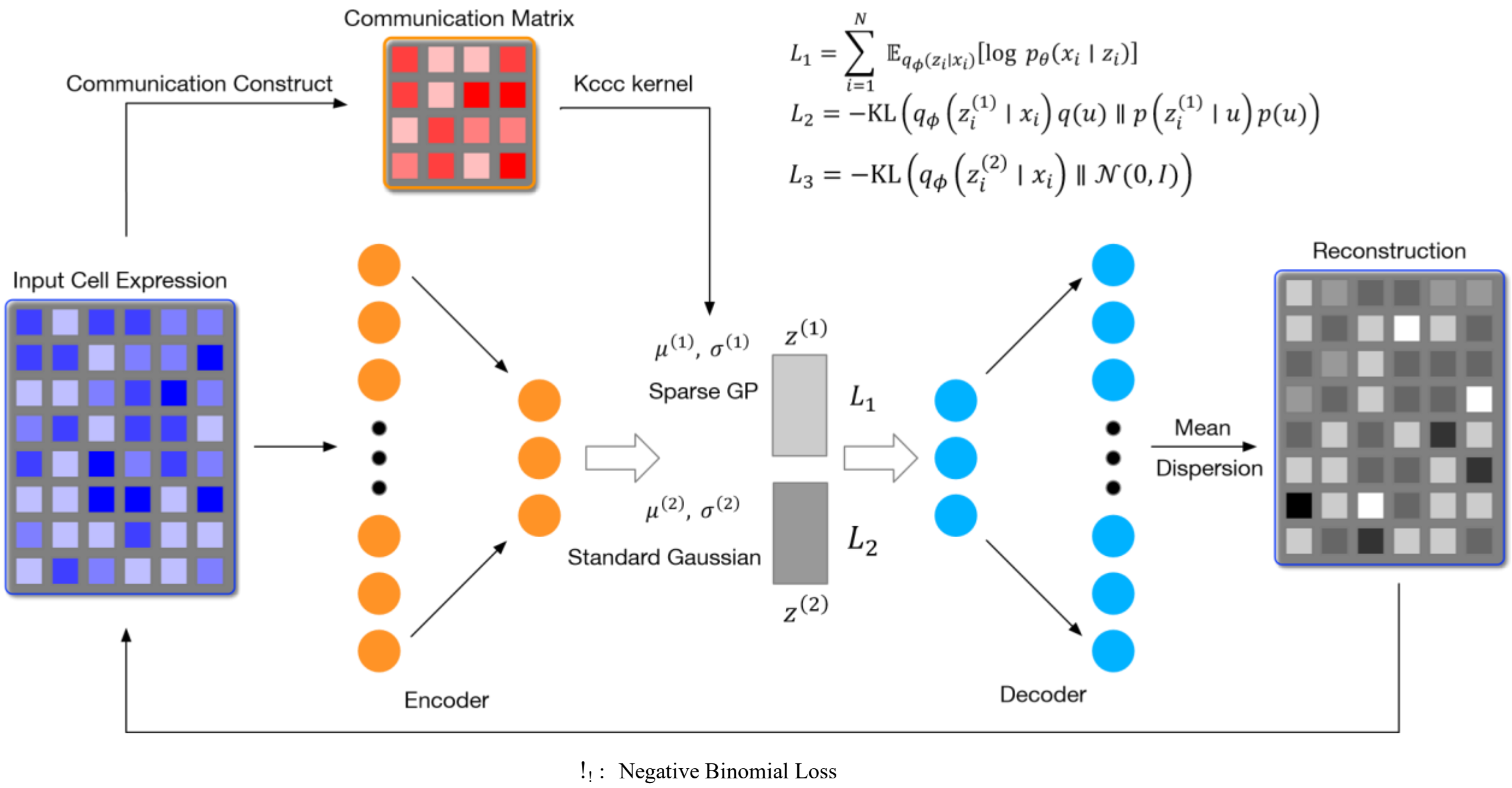

Figure 2: CCCVAE integrates cell-cell communication (CCC) into a variational autoencoder using a Gaussian Process informed by ligand-receptor interactions. Starting from a gene expression matrix, a communication matrix is computed via CellChat to derive the CCC kernel. The encoder maps cells into latent space, where the first $l$ dimensions are shaped by the CCC kernel and the rest model non-CCC variation. Latent vectors are decoded back to gene expression using a Negative Binomial loss.

## 3.2 Constructing the CCC kernel

The construction of the Cell-Cell Communication (CCC) kernel is central to our model. It incorporates biological communication between cells and ensures positive-definiteness for the Gaussian Process. The kernel construction follows these steps:

Step 1: Calculating the communication matrix

We start by calculating the communication probability between ligand-receptor pairs, leveraging a detailed formula from the CellChat [18] framework. This formula models the interaction strength between cells based on gene expression data. For a given ligand $L_i$ in cell $i$ and receptor $R_j$ in cell $j$, the communication probability $P_{ij}$ is calculated as:

$$P_{ij} = \frac{L_i \cdot R_j}{K_h + L_i \cdot R_j} \cdot \left(1 + \frac{AG_i}{K_h + AG_i}\right) \cdot \left(1 + \frac{AG_j}{K_h + AG_j}\right) \cdot \frac{K_h}{K_h + AN_i} \cdot \frac{K_h}{K_h + AN_j} \cdot \frac{1 + RA_j}{1 + RI_j} \tag{2}$$

Here:

- $L_i$ and $R_j$ are the expression levels of ligand $L$ and receptor $R$ for cells $i$ and $j$, respectively.
- $AG_i, AG_j$ represent the soluble agonists for cells $i$ and $j$, respectively.
- $AN_i, AN_j$ represent antagonists for cells $i$ and $j$.
- $RA_j, RI_j$ represent stimulatory and inhibitory receptors in cell $j$.
- $K_h$ is a constant (typically set to 0.5) for normalization.

This equation ensures that the ligand-receptor interaction is modulated by multiple cofactors, including agonists, antagonists, and receptor activities. The geometric mean is used to combine the expressions of multi-subunit ligands and receptors.

Step 2: Incorporating the communication matrix into the kernel function

The Cauchy kernel is used as the base kernel function, which computes the similarity between cells based on their gene expression. The Cauchy kernel is defined as:

$$K_{\text{Cauchy}}\left(x_i, x_j\right) = \frac{1}{1+\frac{d\left(x_i, x_j\right)}{s}} \tag{3}$$

where $s$ is a scale parameter controlling the kernel width (fixed or learnable). The squared distance between cell feature vectors $x_i$ and $x_j$ is computed as

$$d(x_i, x_j) = \|x_i\|^2 + \|x_j\|^2 - 2\langle x_i, x_j\rangle. \tag{4}$$

The communication matrix $C$ is then integrated into the kernel by applying a propagation matrix $P$ that reflects the cell-cell communication. The propagation matrix is defined as:

$$P = I + \beta \cdot C \tag{5}$$

where: $I$ is the identity matrix. $\beta$ is a hyperparameter controlling the strength of the communication effect.

The final kernel function, incorporating both the Cauchy kernel and the communication information, is then computed as:

$$K_{CCC} = K_{\text{Cauchy}} \cdot P \cdot K_{\text{Cauchy}T} \tag{6}$$

This kernel captures the similarity between test and inducing cells while integrating the cell-cell communication information through the propagation matrix $P$ . Details on the positive-definiteness of $K_{CCC}$ are provided in Appendix A.

## 3.3 Sparse GP prior for CCC dimensions

We place an independent sparse variational Gaussian Process prior over each CCC-aware latent dimension $z_i^{(1)}$ . Given inducing points $Z_u \in \mathrm{R}^{m\times G}$ and corresponding inducing outputs $u \in \mathrm{R}^{m\times\updownarrow}$, the approximate posterior for each dimension $l$ is:

$$q\left(f_l\left(x_i\right)\right) = \mathcal{N}\left(\mu_l\left(x_i\right), \sigma_l^2\left(x_i\right)\right) \tag{7}$$

where:

$$
\begin{aligned}
\mu_l(x_i) &= \frac{N}{b} K_{x_i m} \left( K_{mm} + \frac{N}{b} K_{mn} \ \text{diag} (\sigma^{-2}) K_{nm} \right)^{-1} K_{mn} \ \text{diag}(\sigma^{-2}) y \\
\sigma_l^2(x_i) &= K_{x_i x_i} - K_{x_i m} K_{mm}^{-1} K_{m x_i} \\
&\quad + K_{x_i m} \left( K_{mm} + \frac{N}{b} K_{mn} \ \text{diag} (\sigma^{-2}) K_{nm} \right)^{-1} K_{m x_i}
\end{aligned} \tag{8}
$$

Each CCC latent dimension maintains variational parameters ( $\hat{\mu}_l, \hat{A}_l$ ) approximating $q(u_l) = \mathcal{N}(\mu_{\hat{l}}, A_{\hat{l}})$. Across all CCC dimensions:

$$
p(Z_{(1)} \mid U_{(1)}) = \prod_{l=1}^{\ell} p(z_{:(1),l} \mid u_l) \tag{9}
$$

We place the detailed sparse GP prior demonstration in the Appendix B.

### 3.4 Standard Gaussian Prior for independent dimensions

For the independent latent variables $z_i^{(2)}$, we place a standard isotropic Gaussian prior:

$$
p\left(z_i^{(2)}\right) = \mathcal{N}(0, I) \tag{10}
$$

The variational posterior for $z_i^{(2)}$ is parameterized by the encoder network as:

$$
q_\phi\left(z_i^{(2)} \mid x_i\right) = \mathcal{N}\left(\mu_i^{(2)}, diag\left(\left(\sigma_i^{(2)}\right)^2\right)\right) \tag{11}
$$

This structure follows the standard VAE formulation.

### 3.5 Variational Inference and Objective function

The overall variational approximation factorizes as:

$$
q_\phi(z_i \mid x_i) = q_\phi\left(z_i^{(1)} \mid x_i\right) q_\phi\left(z_i^{(2)} \mid x_i\right) \tag{12}
$$

and the posterior over the inducing variables:

$$
q(u) = \mathcal{N}(\mu_u, \Sigma_u) \tag{13}
$$

The Evidence Lower Bound (ELBO) for CCCVAE is:

$$
\begin{aligned}
\mathcal{L}_{CCCVAE} = \sum_{i=1}^{N} & \mathbb{E}_{q_\phi(z_i|x_i)} [\log \text{NB}\,(x_i \mid \mu_\theta(z_i), \theta)] \\
& - KL\left(q_\phi\left(z_i^{(1)} \mid x_i\right) q(u) \,\|\, p\left(z_i^{(1)} \mid u\right) p(u)\right) \\
& - KL\left(q_\phi\left(z_i^{(2)} \mid x_i\right) \,\|\, \mathcal{N}(0, I)\right)
\end{aligned} \tag{14}
$$

This loss function optimizes the variational autoencoder by maximizing the expected log-likelihood of the observed gene expression counts under a Negative Binomial (NB) decoder, which is well-suited for modeling overdispersed scRNA-seq data. [35, 23, 9] It also includes two KL divergence regularization terms: one that aligns the CCC-aware latent dimensions $z_i^{(1)}$ with a Gaussian Process prior mediated by inducing variables $u$, and another that regularizes the remaining latent dimensions $z_i^{(2)}$ with a standard isotropic Gaussian prior to capture residual variation not explained by cell-cell communication.

## 4 Experiments

We evaluate our method on four public single-cell datasets: Opium [41], Pancreas [12], PBMC4K [11], and PBMC12K [13], and compare it with baseline models including Vanilla VAE, GMVAE, scVI, and CCCVAE. As this is an unsupervised task, all cells are used for training. Code and data are available at https://github.com/ Anonymous324-star/CCCVAE.

### 4.1 Experimental setup

For all datasets, we apply the same preprocessing pipeline[43, 36, 4], filtering rare genes and cells, and retaining the top 10,000 highly variable genes. Ligand-receptor interactions are identified using the "consensus" database. Model training is performed using the Adam[19] optimizer with an initial learning rate of $1 \times 10^{-3}$ and a weight decay of $1 \times 10^{-6}$. Each model is trained for up to 100 iterations, with a KL warm-up schedule starting from $\beta$ = 0.1. The reconstruction loss is based on the Negative Binomial (NB) loss function. All experiments are repeated using five different random seeds, and average performance is reported. We use Adjusted Rand Index (ARI) [17] and Normalized Mutual Information (NMI) [32] as the primary evaluation metrics.

### 4.2 CCC improvement analysis

To specifically evaluate the extent to which models preserve Cell-Cell Communication (CCC) structure in the latent space, we conduct a CCC-based evaluation using the PBMC4K and Pancreas datasets. We first use scVI and our proposed CCCVAE to generate latent representations of all cells and apply K-means clustering to assign cluster labels. Next, we compute group-level communication matrices via the CellChat framework, resulting in an adjacency matrix of size $n$ _cluster $\times n$ _cluster, where each entry reflects the communication strength between two cell clusters.

To interpret the preserved signaling structure, we analyze and visualize the resulting communication matrices by plotting the distribution of edge weights, shown as both histograms and cumulative distribution functions (CDFs) in Figure 3. Each subplot compares the edge weight distributions obtained from CCCVAE (blue) and scVI (orange).

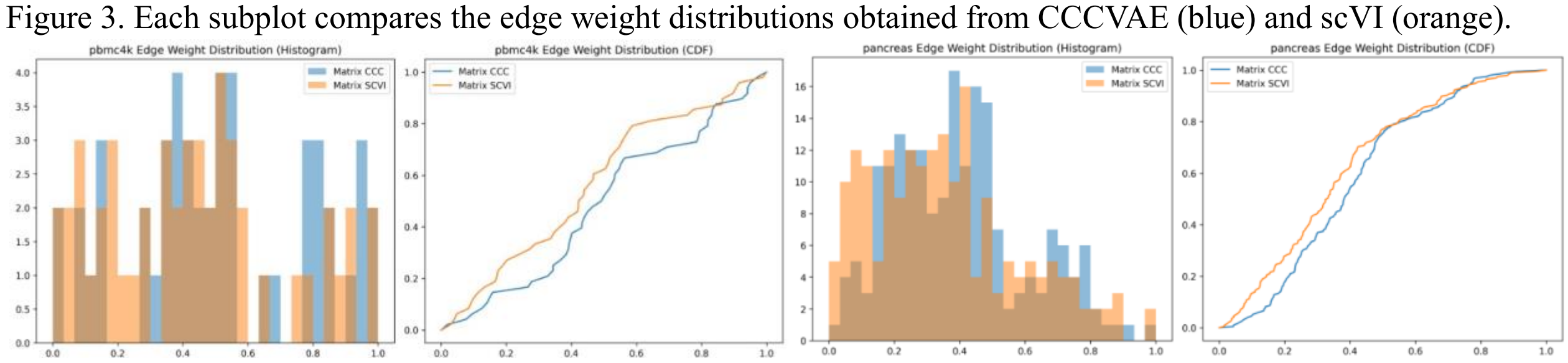

Figure 3: Comparison of edge weight distributions from cell-cell communication graphs derived using latent representations from scVI and CCCVAE on the PBMC4K and Pancreas datasets. Histograms (left) and CDFs (right) show that CCCVAE produces more variable and stronger edge weights, indicating enhanced preservation of inter-cluster signaling structure in the latent space.

Across both datasets, several key trends emerge:

Higher Edge Weight Variability:

CCCVAE yields a broader spread of edge weights, as seen in the histograms, suggesting that it better differentiates between strong and weak inter-cluster communications. In contrast, the distributions from scVI are more concentrated around mid-range values, indicating a lack of contrast in inferred communication strengths.

Right-Shifted CDFs:

The CDF curves for CCCVAE are consistently right-shifted compared to those from scVI , meaning a larger proportion of edges have higher weights. This shift indicates that CCCVAE captures more intense signaling relationships between cell groups.

Dataset Consistency:

These trends hold across both PBMC4K and Pancreas datasets, showcasing the robustness of CCCVAE in different biological contexts. Notably, for the pancreas dataset, the histogram shows a clear enhancement in edge frequency across medium-to-high weight ranges, reinforcing the model's ability to recover biologically plausible signaling pathways.

These findings support our hypothesis that explicitly modeling cell-cell communication during training promotes the emergence of functionally coherent latent structures. By embedding communication priors into the generative process, CCCVAE encourages the latent space to reflect not just transcriptomic similarity, but also functional interaction patterns. This leads to improved interpretability and utility in downstream tasks such as pathway analysis, cell type interaction mapping, and therapeutic target discovery.

## 4.3 Clustering results

We evaluate the clustering performance of CCCVAE and three baseline models-Vanilla VAE, GMVAE, and scVI-across four diverse single-cell datasets: PBMC4K, Opium, Pancreas, and PBMC12K. Clustering quality is assessed using two standard metrics: Adjusted Rand Index (ARI) and Normalized Mutual Information (NMI). The results are summarized in Table 1. CCCVAE consistently achieves the highest ARI and NMI scores across all datasets. These gains are particularly evident on Pancreas and PBMC12K, which are more biologically complex and present greater challenges for cell-type separation. For example, on the Pancreas dataset: ARI improves from 0.177 (Vanilla VAE) and 0.227 (scVI) to 0.287 (CCCVAE), NMI improves from 0.295 (Vanilla VAE) and 0.436 (scVI) to 0.480 (CCCVAE). Similarly, on the PBMC12K dataset: CCCVAE reaches an ARI of 0.462, surpassing scVI's 0.460 and significantly outperforming GMVAE (0.222), CCCVAE attains the highest NMI of 0.670, exceeding scVI's 0.646 and Vanilla VAE's 0.567. These improvements highlight CCCVAE's ability to model cell-type-specific structures with higher biological fidelity. To further support these quantitative results, we present UMAP [26] visualizations of the latent space for the Opium dataset (Figure 4). Compared to scVI, which shows more diffuse and overlapping clusters, CCCVAE produces well-separated and compact clusters that align closely with known cell types. This reflects its improved representation of the intrinsic biological structure.

We attribute CCCVAE’s superior clustering performance to its novel integration of CellCell Communication (CCC) priors through a sparse Gaussian Process layer. These priors encode biologically informed interactionssuch as ligand-receptor signaling-into the latent space. As a result, CCCVAE captures not only individual expression patterns but also relational dependencies that guide cell differentiation and spatial organization. This biologically grounded structure encourages the emergence of more coherent, interaction-aware clusters, setting CCCVAE apart from existing unsupervised models.

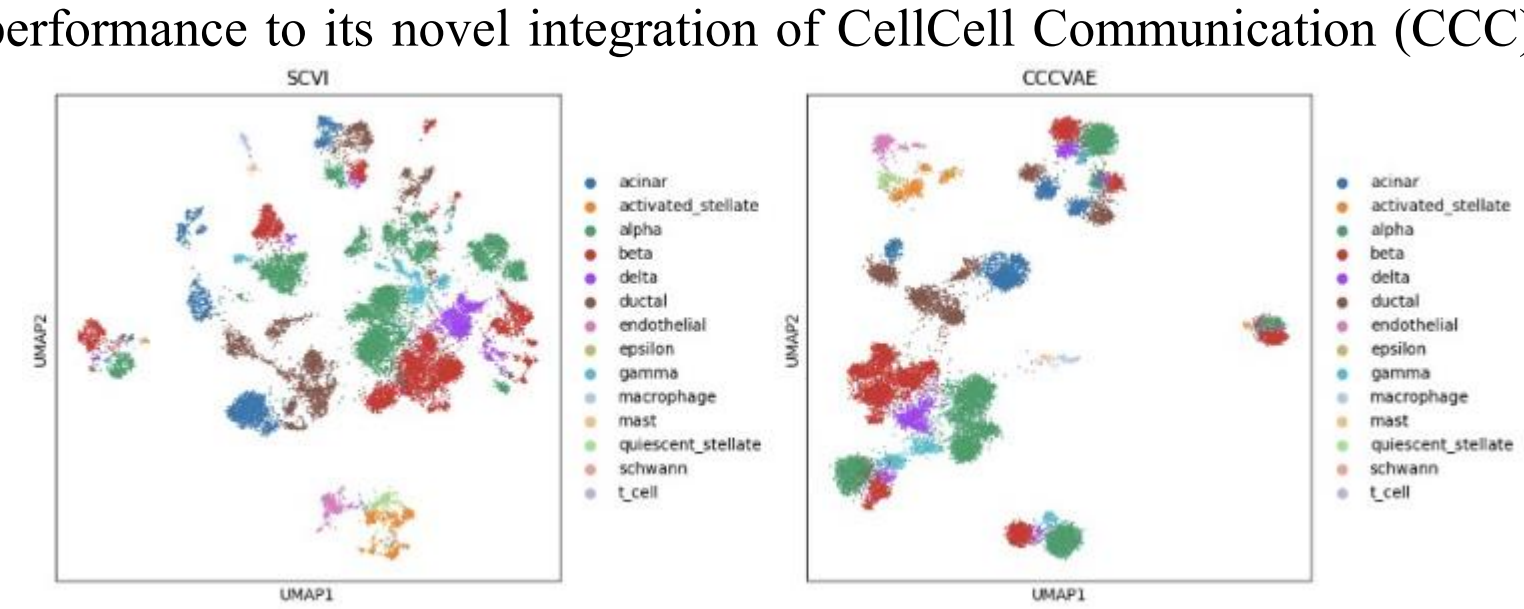

Figure 4: The visualization of the clustering result on PBMC12K dataset.

Table 1: Clustering results of Vanilla VAE, GMVAE, scVI, and CCCVAE on Four Datasets

| Model | PBMC4K | | Opium | | Pancreas | | PBMC12K | |
|---|---|---|---|---|---|---|---|---|
| | ARI | NMI | ARI | NMI | ARI | NMI | ARI | NMI |
| Vanilla VAE | 0.402 | 0.259 | 0.119 | 0.332 | 0.177 | 0.295 | 0.342 | 0.567 |
| GMVAE | 0.015 | 0.121 | 0.179 | 0.214 | 0.028 | 0.152 | 0.222 | 0.218 |
| scVI | 0.403 | 0.610 | 0.387 | 0.670 | 0.227 | 0.436 | 0.460 | 0.646 |
| CCCVAE | 0.450 | 0.605 | 0.418 | 0.703 | 0.287 | 0.480 | 0.462 | 0.670 |

## 4.4 Biological Interpretability

To rigorously assess the biological coherence of the cell populations inferred by CCCVAE, we conducted an indepth characterization of marker gene expression and their associated functional programs across clusters. This analysis aimed to determine whether the transcriptomic patterns captured in the latent space reflect biologically meaningful cellular identities and functions.

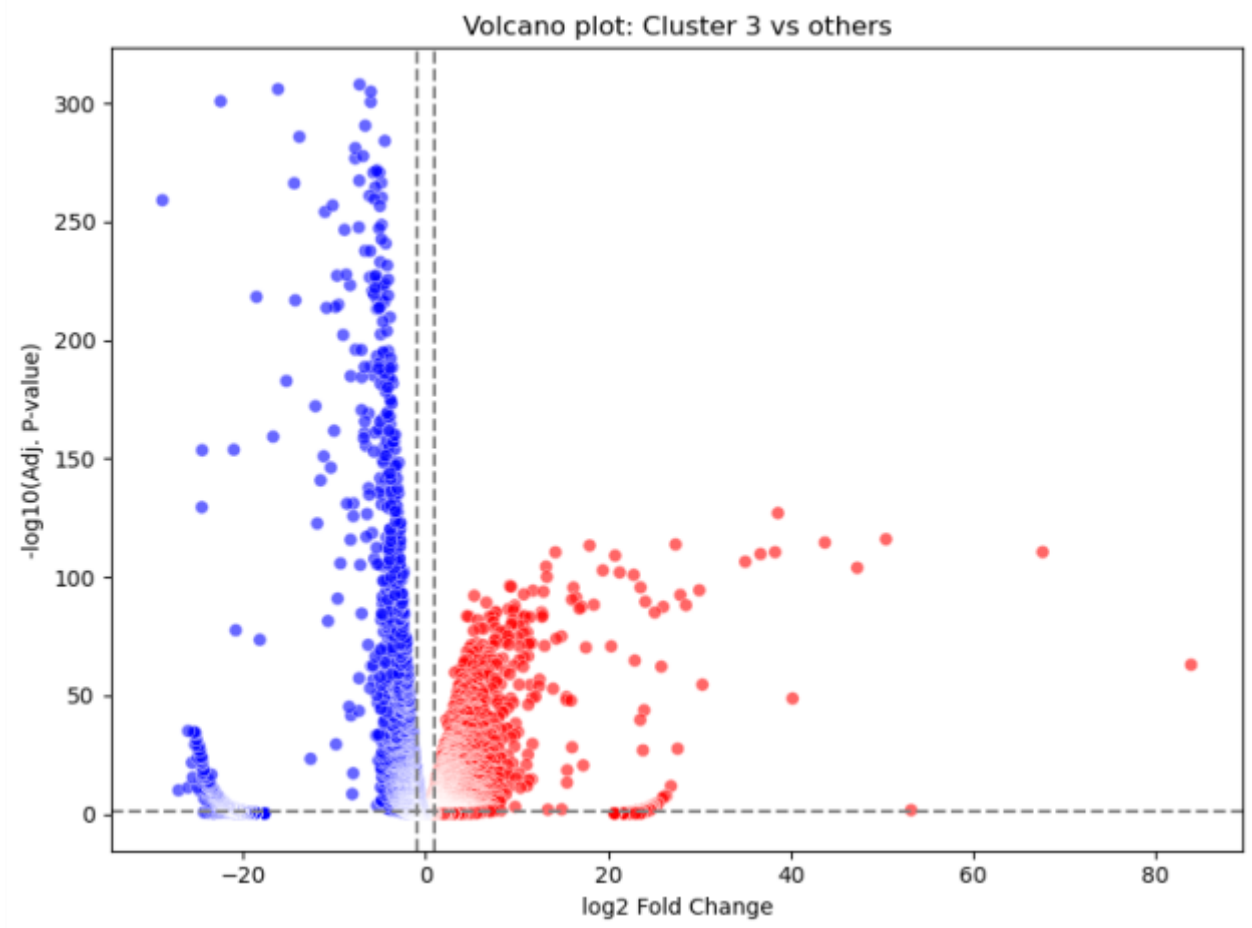

Figure 5: Volcano plot of differential gene expression comparing cluster 3 versus all other clusters. Each point represents a gene, with $\log_2$ fold change on the x-axis and $\log_{10}$ adjusted p-value on the y-axis. Red and blue points indicate significantly up- and down-regulated genes, respectively ($\log_2$FC > 1 and adjusted $p < 0.05$).

Marker gene expression patterns. We first visualized the expression of the top 20 differentially expressed genes (DEGs) across all clusters using a scaled heatmap (Figure 5). This visualization highlights highly structured transcriptional landscapes, with distinct clusters exhibiting mutually exclusive or sharply enriched gene signatures. For example, cluster 0 shows strong expression of canonical oligodendrocyte markers such as *PLP1*, *MBP*, and *MOG*, consistent with a myelinating glial phenotype crucial for axonal insulation in the central nervous system. In contrast, cluster 3 displays high expression of neuronal markers, including *RBFOX3* (NeuN) and *SYT1*, which are essential for synaptic vesicle trafficking and neuronal excitability. Other clusters demonstrate characteristic signatures as well; for instance, clusters 1 and 4 display low expression of neuronal markers but are enriched for glial or support-cell-related transcripts. Together, these patterns suggest that CCCVAE successfully recovers cell populations that align with canonical neural lineages.

Differential expression analysis of neuronal clusters. To more systematically evaluate transcriptional specificity, we performed a differential expression (DE) analysis comparing cluster 3 against all other clusters. For each gene, we computed $\log_2$ fold change to capture the magnitude of regulation and used Welch's t-test to assess statistical significance, followed by Benjamini–Hochberg correction to control for multiple testing. The volcano plot in Figure 5 reveals a broad distribution of $\log_2$ fold changes, with many neuronal genes showing both large effect sizes and extremely low adjusted p-values. Genes passing the threshold of $\log_2$FC > 1 and adjusted $p < 0.05$ are highlighted in red (upregulated) and blue (downregulated). Notably, cluster 3 shows significant enrichment for genes involved in synaptic function, excitatory neurotransmission, and calcium signaling, underscoring its neuronal identity.

Functional enrichment and biological interpretation. To elucidate the functional roles of the genes driving the identity of cluster 3, we conducted Gene Ontology (GO) enrichment analysis within the Biological Process category. Upregulated genes were strongly enriched for processes related to excitatory synaptic transmission, including *glutamate receptor signaling pathway*, *modulation of chemical synaptic transmission*, and *regulation of NMDA receptor activity* (Figure 6). Many of these pathways are central to neuronal communication, plasticity, and circuit integration, suggesting that cluster 3 represents a population of mature, synaptically active excitatory neurons. Bubble size in the enrichment plot reflects the number of genes associated with each term, while color represents $\log_{10}$FDR, highlighting the most statistically significant biological processes.

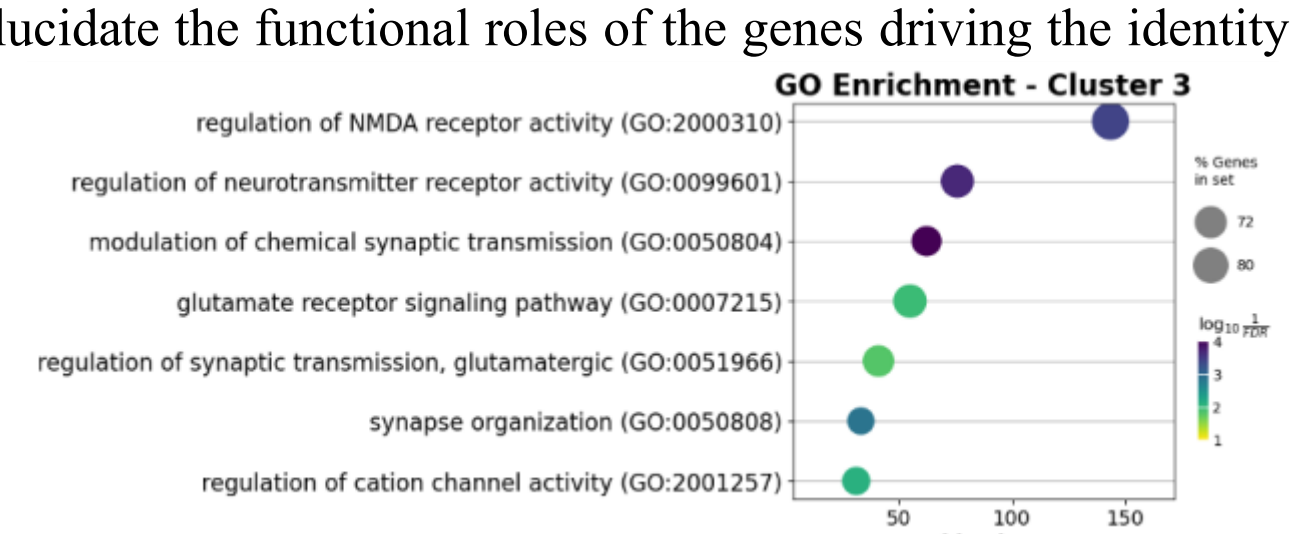

Figure 6: GO enrichment bubble plot for upregulated genes in cluster 3, showing enriched biological processes related to synaptic signaling and excitatory neurotransmission. Bubble size reflects gene count; color indicates $-\log_{10}$(FDR).

## 5 Ablation Study

To comprehensively understand the role and sensitivity of key components within our proposed model, we performed a systematic ablation study focusing on two critical hyperparameters: the number of latent dimensions allocated to the Gaussian Process (GP) component and the KL-divergence regularization coefficient $\beta$.

Gaussian Process Dimension The first ablation focuses on the number of latent dimensions dedicated to the GP component, which governs how much of the latent space is explicitly structured by cell-cell communication signals. We fixed the total latent dimension to 80 to ensure a fair comparison, and varied the number of dimensions controlled by the GP kernel to 4, 8, 16, and 32. The remaining dimensions in each configuration were modeled using a standard isotropic Gaussian, enabling us to isolate the impact of the GP-structured subspace.

As illustrated in Figure 7a, the Adjusted Rand Index (ARI) scores consistently improve as the number of GP dimensions increases from 4 to 16 across all four benchmark datasets (PBMC4K, Opium, Pancreas, and PBMC12K). This trend suggests that increasing the GP capacity allows the model to better capture structured dependencies arising from biological signaling. Notably, the performance reaches its peak at 16 GP dimensions. Beyond this point, a further increase to 32 dimensions leads to a noticeable decline in ARI, likely due to overfitting or the dilution of general-purpose latent structure. These results highlight a critical trade-off: too few GP dimensions may underutilize the model's capacity to encode communication-driven variability, while too many can harm generalization by excessively constraining the representation. The 16-dimensional setting emerges as a sweet spot, providing sufficient flexibility for communication modeling without undermining other latent patterns.

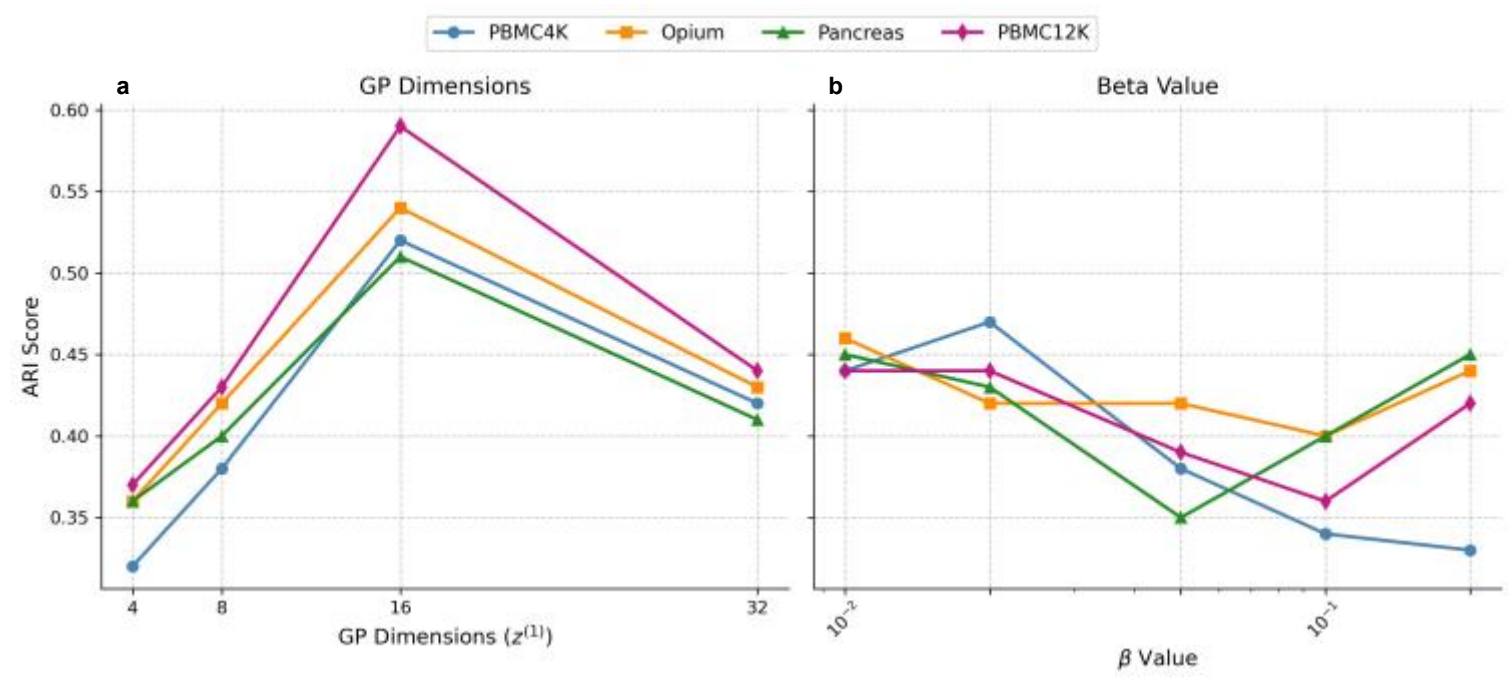

Figure 7: Ablation Study of GP Dimensions and Beta Value

KL-Divergence Weight

The second ablation investigates the role of the KL-divergence coefficient $\beta$, which scales the regularization term in the VAE loss function and determines how strongly the approximate posterior is encouraged to match the prior distribution. In biological contexts, over-regularization can suppress meaningful cell-to-cell variation, while under-regularization may lead to overfitting. We evaluated five values: $[0.01, 0.02, 0.05, 0.1, 0.2]$, and examined their effect on clustering performance. As shown in Figure 7b, smaller $\beta$ values (particularly 0.01 and 0.02) lead to higher ARI scores, consistently outperforming larger $\beta$ settings. This trend indicates that a looser constraint on the posterior enables the model to retain more nuanced and heterogeneous features in the latent space, which are crucial for accurately delineating cell types. Conversely, higher $\beta$ values (0.1 and 0.2) impose stricter adherence to the prior, which can prematurely collapse the latent structure, reducing its utility for downstream clustering tasks. These results reaffirm the need for careful calibration of the KL weight to avoid overly simplistic representations while preserving the diversity of biological signals.

## 6 Conclusion

In this work, we presented CCCVAE, a novel variational framework that integrates cell-cell communication (CCC) signals into the latent space of single-cell data through a sparse Gaussian Process prior. By embedding biologically meaningful interactions directly into representation learning, CCCVAE improves clustering accuracy and enhances the interpretability of latent structures. Empirical results across multiple datasets confirm its consistent performance gains over baseline models. CCCVAE bridges the gap between representation learning and intercellular communication analysis, offering new possibilities for identifying signaling pathways, guiding therapeutic target discovery, and refining cell type annotations. While CCCVAE improves latent space structure, its performance still depends on the quality of the ligand–receptor interaction database and the accuracy of inferred communication matrices. Future work could explore integrating additional biological priors, such as spatial transcriptomics or multi-omics data, to further refine latent representations and enhance clustering accuracy.

# Appendix

## A Positive-Definiteness of the Kernel

The final kernel $K_{CCC}$ remains positive-definite because:

- The Cauchy kernel $K_{\text{Cauchy}}$ is positive-definite by design, as it represents a valid covariance function for a Gaussian Process.
- The propagation matrix $P$ is constructed using the communication matrix, which is symmetric and scaled by the identity matrix. As $P$ is derived from the communication matrix, which is positive semidefinite, and it is modified by adding the identity matrix, it remains positive semi-definite.
- The product of two positive-definite matrices (i.e., $K_{\text{Cauchy}}$ and $P$ ) preserves positive-definiteness, ensuring the overall kernel is valid for Gaussian Process inference.

Thus, this kernel construction ensures both biological relevance (cell communication is captured) and computational validity (positive-definiteness for inference).

*Proof*

To formally prove that $K_{CCC} = K_{\text{Cauchy}} \cdot P \cdot K_{\text{Cauchy}}^T$ is positive definite, given that $K_{\text{Cauchy}}$ and $P$ are positive definite, we need to show that for any non-zero vector $v$, the quadratic form $v^T K_{CCC} v$ is strictly positive.

1. Define the quadratic form for $K_{CCC}$:

We start with the quadratic form $v^T K_{CCC} v$, where $K_{CCC} = K_{\text{Cauchy}} \cdot P \cdot K_{\text{Cauchy}}^T$ .

$$v^T K_{CCC} v = v^T K_{\text{Cauchy}} \cdot P \cdot K_{\text{Cauchy}}^T )\, v$$

2. Rearrange the expression:

We can rewrite the quadratic form by associating the matrices:

$$v^T K_{CCC} v = \left(K_{\text{Cauchy}}^T v\right)^T P \left(K_{\text{Cauchy}}^T v\right)$$

This is equivalent to:

$$v^T K_{CCC} v = z^T P z$$

where $z = K_{\text{Cauchy}}^T v$.

3. Positive Definiteness of $P$ :

Since $P$ is positive definite, we know that for any non-zero vector $z, z^T P z > 0$. This follows directly from the definition of a positive definite matrix.

4. Non-zero vector $z$ :

The vector $z = K_{\text{Cauchy}}^T v$ is non-zero for any non-zero $v$, because $K_{\text{Cauchy}}$ is assumed to be positive definite. A positive definite matrix does not map non-zero vectors to the zero vector, so $z \neq 0$ for $v \neq 0$.

5. Conclusion:

Since $P$ is positive definite, for any non-zero vector $z$, we have:

$$z^T P z > 0$$

Therefore, the quadratic form $v^T K_{CCC} v$ is strictly positive for any non-zero vector $v$, which implies that $K_{CCC}$ is positive definite.

## B Sparse GP Prior Correctness

We provide the detailed derivation showing that our sparse variational Gaussian Process (GP) prior correctly approximates the posterior over the CCC-aware latent dimensions $z_i^{(1)}$.

### B.1 Model assumption

For each CCC-aware latent dimension $l \in \{1,...,\updownarrow\}$ :

- $f_l(x) \sim \mathcal{GP}\left(0, K_{CCC}\left(x, x'\right)\right)$ (Zero-mean GP with CCC kernel.)
- Inducing points $Z_u = [u_1,...,u_m] \in \mathbb{R}^{m\times G}$ with outputs $u_l \in \mathbb{R}^m$.
- Given observed noisy data $(x_i, y_i)$, we assume:

$$y_i = f_l\left(x_i\right) + \epsilon_i,\ \epsilon_i \sim \mathcal{N}\left(0, \sigma_i^2\right)$$

where $\sigma_i^2$ is the encoder-predicted noise variance.

- Variational distribution over inducing outputs:

$$q\left(u_l\right) = \mathcal{N}\left(\mu_{\hat{l}}, A_{\hat{l}}\right)$$

Variational Approximation of posterior

Given the standard variational GP framework (Titsias, 2009), the exact posterior is intractable. We introduce $q(u_l)$ and derive an approximate posterior $q(f_l(x))$ as follows:

The joint prior over $f_l(X)$ and $u_l$ is:

$$p(f_l(X), u_l) = \mathcal{N}\left(0, \begin{bmatrix} K_{nn} & K_{nm} \\ K_{mn} & K_{mm} \end{bmatrix}\right)$$

where:

- $K_{nn}$ is the kernel between training points.
- $K_{nm}, K_{mn}$ are cross-covariances.
- $K_{mm}$ is the kernel among inducing points.

Using conditional properties of Gaussian distributions:

$$p(f_l(X) \mid u_l) = \mathcal{N}\left(K_{nm}K_{mm}^{-1}u_l,\ K_{nn} - K_{nm}K_{mm}^{-1}K_{mn}\right)$$

Thus, the variational posterior becomes:

Z $q(f_l(X)) =$ $p(f_l(X)$

$\mid u_l)q(u_l)du_l$

Since $p(f_l(X) \mid u_l)$ and $q(u_l)$ are both Gaussian, $q(f_l(X))$ remains Gaussian, and its mean and covariance are:

$$\mathbb{E}_{q(u_l)}\left[f_l(x)\right] = K_{nm}K_{mm}^{-1}\mu_{\hat{l}}$$
$$Cov_{q(u_l)}\left[f_l(x)\right] = K_{nn} - K_{nm}K_{mm}^{-1}K_{mn} + K_{nm}K_{mm}^{-1}A_iK_{mm}^{-1}K_{mn}$$

B.2 Minibatch Correction

Since we optimize using minibatches, following (Hensman et al., 2013), we rescale the noise by $\frac{N}{b}$ :

- Scaled noise variance:$\widetilde{\sigma}_i^2 = \frac{b}{N}\sigma_i^2$.

The corrected variational posterior for minibatch inference is:

$$q\left(f_l\left(x_i\right)\right) = \mathcal{N}\left(\mu_l\left(x_i\right), \sigma_l^2\left(x_i\right)\right)$$

with:

$$\mu_l\left(x_i\right) = \tfrac{N}{b}K_{x_im}\left(K_{mm} + \tfrac{N}{b}K_{mn}diag\left(\sigma^{-2}\right)K_{nm}\right)^{-1}K_{mn}diag\left(\sigma^{-2}\right)y$$
$$\sigma_l^2\left(x_i\right) = K_{x_ix_i} - K_{x_im}K_{mm}^{-1}K_{mx_i} + K_{x_im}\left(K_{mm} + \tfrac{N}{b}K_{mn}diag\left(\sigma^{-2}\right)K_{nm}\right)^{-1}K_{mx_i}$$

These are exactly the formulas implemented in our method.

B.3 Variational Evidence Lower Bound (ELBO) The

sparse GP ELBO for each CCC-aware dimension $l$ is:

$$\mathcal{L}_{GP}^{(l)} = \mathbb{E}_{q(f_l(x))}\left[logp\left(y \mid f_l(x)\right)\right] - KL\left(q\left(u_l\right) \parallel p\left(u_l\right)\right)$$

Explicitly:

- The likelihood term includes corrections for the minibatch noise.
- The KL divergence is:

$$KL\left(q\left(u_l\right) \parallel p\left(u_l\right)\right) = \frac{1}{2}\left(log\frac{detK_{mm}}{detA_i} - m + Tr\left(K_{mm}^{-1}A_i\right) + \mu_i^\top K_{mm}^{-1}\mu_i\right)$$

Thus, the variational lower bound is minimized correctly through minibatch-optimized ELBO.

## C Datasets and Baselines

Opium: A subset of a ventral midbrain dataset profiling single nuclei from opioid overdose cases and drug-free controls, capturing glial immune activation and neuronal synaptic repression. We use the 'M16' matrix in this study.

Pancreas: A pancreas cell dataset collected from multiple donors, exhibiting strong batch effects and diverse cell types.

PBMC4K: Peripheral blood mononuclear cells dataset with approximately 4,000 cells.

PBMC12K: A larger PBMC dataset with around 12,000 cells, used to evaluate scalability.

Table 2: Statistics of the evaluated single-cell RNA-seq datasets.

| Dataset | #Cells | #Genes | #Cell Types |
|---|---|---|---|
| Opium | 5,437 | 36,601 | 9 |
| Pancreas | 16,382 | 19,093 | 14 |
| PBMC4K | 3,617 | 33,538 | 7 |
| PBMC12K | 11,990 | 3,346 | 9 |

We compare CCCVAE model's clusterness performance against the baseline models:

Vanilla VAE: A standard Variational Autoencoder with isotropic Gaussian prior and simple reconstruction loss.

Gaussian Mixture VAE (GMVAE): A VAE variant with a Gaussian Mixture prior to better model clustering structures. scVI: A state-of-the-art deep generative model specifically designed for single-cell data, incorporating a Negative Binomial reconstruction loss and batch effect correction.

CCCVAE: Our proposed model, which integrates Cell-Cell Communication (CCC) information into the latent space via a sparse Gaussian Process prior.

Setup details

All experiments are conducted on a Linux machine (Ubuntu 16.04.4 LTS (GNU/Linux 4.4.0-210-generic x86_64)) equipped with a single NVIDIA A100 GPU (80 GB) and 128 GB of RAM. The CUDA version is 12.4 and the NVIDIA driver version is 550.90.12. All code is implemented in Python 3.9.13 using PyTorch 2.6.0.